\pdfoutput=1
\documentclass[letterpaper, 10 pt, journal, twoside]{IEEEtran} 

\IEEEoverridecommandlockouts                              

\usepackage{amsthm}
\usepackage{xfrac}
\usepackage{nicefrac}
\usepackage{booktabs}
\usepackage{flushend}
\usepackage[switch, pagewise]{lineno}
\usepackage{mathrsfs}
\DeclareMathAlphabet{\mathpzc}{OT1}{pzc}{m}{it}
\usepackage{multirow}
\usepackage{multicol}
\usepackage{bbm}
\usepackage{soul}
\usepackage{xfrac}
\usepackage{balance}
\usepackage{titlesec}
\usepackage{caption}
\usepackage{subcaption}
\usepackage[final]{pdfpages}
\usepackage{hyperref}
\usepackage{resizegather}
\usepackage{svg}
\usepackage{ctable}
\usepackage{gensymb}
\usepackage{cite}
\usepackage{xcolor}
\usepackage{wasysym}

\usepackage{amsmath}
\usepackage{amssymb}
\usepackage{mathtools}
\usepackage{graphicx}
\usepackage{float}
\usepackage{array}
\usepackage{listings}
\usepackage{graphicx}
\usepackage{cite}
\usepackage{dsfont}
\usepackage{mathtools,etoolbox}
\usepackage[ruled, linesnumbered]{algorithm2e}
\usepackage[none]{hyphenat}
\usepackage[utf8]{inputenc}
\usepackage[english]{babel}
\usepackage[T1]{fontenc}

\usepackage{amsthm}
\usepackage{xfrac}
\usepackage{nicefrac}
\usepackage{booktabs}
\usepackage{flushend}
\usepackage[switch, pagewise]{lineno}
\usepackage{mathrsfs}
\DeclareMathAlphabet{\mathpzc}{OT1}{pzc}{m}{it}
\usepackage{multirow}
\usepackage{multicol}
\usepackage{bbm}
\usepackage{soul}
\usepackage{xfrac}
\usepackage{balance}
\usepackage{titlesec}
\usepackage{caption}
\usepackage{subcaption}
\usepackage[final]{pdfpages}
\usepackage{hyperref}
\usepackage{resizegather}
\usepackage{svg}
\usepackage{ctable}

\hypersetup{colorlinks=true,urlcolor=blue,citecolor=red, allcolors=blue}
\begin{document}

\makeatletter



\title{\LARGE \bf \textit{AsterNav}: Autonomous Aerial Robot Navigation In Darkness Using Passive Computation} 
%

\author{\textcolor{black}{Deepak Singh}$^{1*}$, Shreyas Khobragade$^{1*}$, Nitin J. Sanket$^{1}$
\thanks{Manuscript received: October, 09, 2025; Accepted December, 15, 2025.}
\thanks{This paper was recommended for publication by Pascal Vasseur upon evaluation of the Associate Editor and Reviewers' comments.
} 
\thanks{$^1$Perception and Autonomous Robotics (PeAR) Group, Worcester Polytechnic Institute. Deepak Singh and Shreyas Khobragade contributed equally, and the author order is decided at random. \textit{(Corresponding
author: Deepak Singh).} Emails: \texttt{\{dsingh1, skhobragade, nitin\}@wpi.edu}.\\
Digital Object Identifier (DOI): see top of this page.}
}

\markboth{PUBLISHED IN IEEE ROBOTICS AND AUTOMATION LETTERS, 2026. DOI: \url{https://doi.org/10.1109/LRA.2026.3653388}}
{Singh and Khobragade \MakeLowercase{\textit{et al.}}: AsterNav}

\maketitle
\begin{abstract}
Autonomous aerial navigation in absolute darkness is crucial for post-disaster search and rescue operations, which often occur from disaster-zone power outages. Yet, due to resource constraints, tiny aerial robots, perfectly suited for these operations, are unable to navigate in the darkness to find survivors safely. In this paper, we present an autonomous aerial robot for navigation in the dark by combining an Infra-Red (IR) monocular camera with a large-aperture coded lens and structured light without external infrastructure like GPS or motion-capture. Our approach obtains depth-dependent defocus cues (each structured light point appears as a pattern that is depth dependent), which acts as a strong prior for our \textit{AsterNet} deep depth estimation model. The model is trained in simulation by generating data using a simple optical model and transfers directly to the real world without any fine-tuning or retraining. \textit{AsterNet} runs onboard the robot at 20 Hz on an NVIDIA Jetson Orin$^\text{TM}$ Nano. Furthermore, our network is robust to changes in the structured light pattern and relative placement of the pattern emitter and IR camera, leading to simplified and cost-effective construction. We successfully evaluate and demonstrate our proposed depth navigation approach \textit{AsterNav} using depth from \textit{AsterNet} in many real-world experiments using only onboard sensing and computation, including dark matte obstacles and thin ropes (\diameter  6.25mm), achieving an overall success rate of \textit{95.5\%} with unknown object shapes, locations and materials. To the best of our knowledge, this is the first work on monocular, structured-light-based quadrotor navigation in absolute darkness.

\end{abstract}


\section*{Supplementary Material}
The accompanying video, supplementary material, code and dataset are available at
\url{http://pear.wpi.edu/research/asternav.html}.


\section{Introduction}
\begin{figure}[t]
    \centering
    \includegraphics[width=\linewidth]{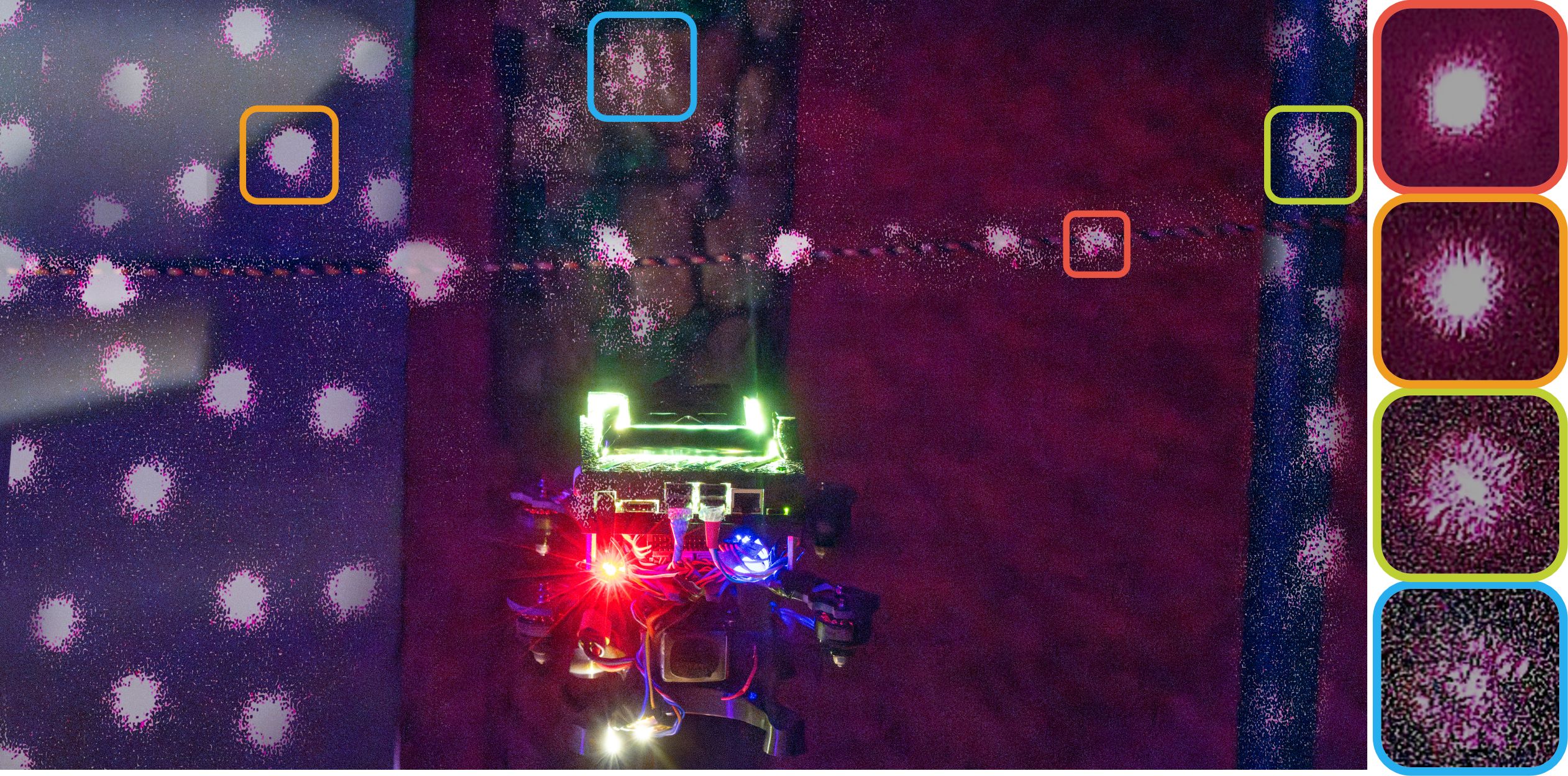}
    \caption{\textit{AsterNav} can estimate dense metric scene depth in complete darkness and navigate around it by using a large-aperture coded lens with a structured lighting source. The system is robust to unknown obstacles of different shapes, sizes, textures, and materials. The highlighted regions on the left illustrate how the projected dot pattern appears at different depths, exhibiting varying degrees of blur \textit{(blue to red denotes increasing depth in highlights)}. \textit{All the images in this paper are best viewed on a computer screen at 200\% zoom at a 100\% brightness.}}
    \label{fig:Banner}
\end{figure}

Darkness is a common occurrence in disaster scenarios that cause damage to infrastructure, power lines and lights. Tiny aerial robots are perfectly suited for disaster search and rescue due to their safety, agility and scalability, yet they are mostly used under daylight conditions. For practical deployment in search and rescue, GPS-denied navigation in the wild, particularly in dark environments such as subterranean tunnels, forests at night, or disaster zones, is crucial and yet remains an open problem. In these environments, vision-based systems often fail due to insufficient illumination. 

Existing autonomy solutions in the dark typically rely on active sensing, such as LIDAR/RADAR or heavy illumination to aid cameras \cite{Barfoot_cam_lidar, radar_cam_fusion, kostas_subt}. Although these approaches work well, they often require tens of Watts of power and weigh a few hundred grams, making them unsuitable for palm-sized aerial robots. Furthermore, high-power lights consuming >10W mounted on aerial robots can restore visibility for conventional monocular or stereo pipelines with algorithmic modifications to account for low signal-to-noise-ratio. Still, they impose severe requirements in terms of power consumption and payload weight, which are critical constraints for tiny aerial robots for extended periods of operation \cite{compra, darpa}. Tight resource constraints on tiny aerial robots make these sensing approaches impractical, forcing reliance on external infrastructure or heavy onboard sensors \cite{nitin_thesis}. Parsimony is the key principle that enables biological flyers like moths to fly in these harsh conditions with remarkable ease \cite{decroon2022insect}. In this work, we propose a novel navigation system that combines monocular vision with structured lighting. Instead of relying on energy-hungry floodlights, our approach projects a sparse, low-power illumination pattern into the environment and interprets its deformation through a monocular camera to infer depth. Inspired by nature, we utilize depth from defocus cues of a large aperture lens \cite{jumping_spider}. \textcolor{black}{We term the class of robot perception as \textit{passive computation}, wherein the input wavefront is modulated \cite{bhandari2022computational} using physical structures that consume no electrical power to inject task-specific informational cues.} 
To improve depth segmentation across distances, we implement passive computation using a coded aperture (similar to those found in Cuttlefish eyes \cite{kim2023cuttlefish}), which produces more distinct and easily distinguishable blur patterns at varying depths. When coupled to a structured light source, we can ``see'' each dot of the dot-pattern to be ``blurred'' out in a specific way that encodes depth at that point. This enables autonomy in complete darkness while maintaining a drastically lower energy footprint. Crucially, our pipeline runs fully onboard a NVIDIA Jetson Orin$^\text{TM}$ Nano at 20Hz, requiring no off-board computation or external localization infrastructure. The system closes the loop from perception to control, achieving fully autonomous navigation in the dark with a palm-sized aerial robot.

\subsection{Problem Formulation and Contributions}
A quadrotor moves in a static scene with obstacles of unknown shape, size, location and material. The quadrotor is equipped with a structured light source and a camera with a coded aperture, and a large-aperture lens. The problem we address is as follows: \textit{Can we present an autonomy framework for the task of static obstacle avoidance-based navigation in complete darkness using only on-board sensing and computation?}

We present both a deep learning and a classical solution for obtaining depth/depth-cues by observing the defocus patterns of the coded aperture, which is then used for obstacle avoidance-based navigation. Our key contributions are:
\begin{itemize}
    \item We propose a novel sensing method to obtain metric depth in darkness by combining passive computation (coded aperture and large aperture lens) and active structured lighting (Fig. \ref{fig:Banner}).
    \item We propose a neural network called \textit{AsterNet} trained only in simulation that generalizes to the real-world without any fine-tuning or re-training for metric depth estimation.
    \item We present a simple classical approach for obstacle avoidance on tiny aerial robots with extreme computational constraints.
    \item We demonstrate and evaluate the proposed navigation approach \textit{AsterNav} on a real quadrotor with on-board sensing and computation in many real-world experiments under various settings, including completely black obstacles and thin ropes in a dark environment.
\end{itemize}

\subsection{Related Work}
\subsubsection{Navigation in low light}
Navigation in low light remains a fundamental challenge, and over the past decade, significant progress has been made with sensor modalities such as LiDAR and RADAR, either individually or in combination with cameras \cite{super, radar_cam_fusion, Barfoot_cam_lidar}. Vision-only approaches are particularly affected in low illumination and have typically relied on external light sources to enable perception \cite{darpa}. \cite{dark_vision} proposed an active vision strategy that adaptively selects seed images and employs camera response modeling to generate HDR-like frames, thereby supporting consistent visual SLAM in environments with brightness levels below 10 lux. Event cameras, owing to their higher dynamic range, have also been explored for robust navigation under low light conditions \cite{bhattacharya2024monoculareventbasedvisionobstacle, ultimateSLAM}. Nevertheless, purely vision-based approaches are not capable of operation in absolute darkness, without relying on high-power external lighting, which is impractical for size and power-constrained aerial robots. To the best of our knowledge, the present work constitutes the first demonstration of monocular frame-based navigation in absolute darkness using a low-power structured lighting source on quadrotors.

\subsubsection{Coded Aperture Imaging and Depth from Defocus}

Nature has evolved parsimonious sensing strategies tailored to the computational limits of organisms, with depth from defocus used by jumping spiders \cite{jumping_spider} and cuttlefish \cite{kim2023cuttlefish}. Adelson and Wang~\cite{adelson1992} extended depth-from-defocus techniques by incorporating a coded mask within a plenoptic camera, thereby enabling the capture of angular information. Levin et al. \cite{coded_aperture_depth_mit} later designed optimized coded apertures that generate depth-dependent blur patterns, enabling reliable joint recovery of image and depth from a single camera. Subsequent works improved efficiency and robustness by enforcing rotational symmetry constraints \cite{metzler} or extending coded designs for computational photography and 3D reconstruction. More recently, learning-based and hardware–software co-design approaches have advanced coded aperture depth sensing, including dual-pixel and split-aperture systems \cite{cads2024,splitaperture2024}, zero-shot phase-coded priors \cite{dpcip2025}, and meta-imaging cameras robust to aberrations \cite{meta2024}. Despite these advances, the use of coded apertures for depth from defocus in robotic navigation remains limited.

\subsection{Organization of the paper} 
\label{subsec:organization}
We first discuss the details of depth from defocus and coded aperture imaging, and then present our data generation approach in \S\ref{sec:asterNet}. This is followed by our \textit{AsterNet} model details, benchtop experiments and real-world flight tests in \S\ref{subsec:model_training} and \S\ref{sec:expts} respectively. Finally, we present a detailed analysis of our results, with parting thoughts on future work in \S\ref{subsec:discussion} and \S\ref{sec:conclusion}. 


\section{\textit{AsterNet}: Dense Depth From Passive Optics with Active Structured Lighting}
\label{sec:asterNet}

Robust obstacle avoidance requires distinguishing near obstacles that demand immediate dodging action from distant structures that can be ignored. Since our approach uses a monocular camera, we will focus the discussion on the same setup. While numerous approaches exist for well-illuminated environments, such as optical flow \cite{ajna, seeing_through_pixel_motion, edgeflownet} or monocular depth \cite{yang2025zeroshotmetricdepthestimation, kim2025careenhancingsafetyvisual, midas}, few methods enable such reasoning in complete darkness. Our approach uses an active structured light mated to a coded aperture lens to provide depth cues through defocus. The coded aperture introduces structured blur patterns whose appearance varies with object distance, thereby enabling depth estimation in dark environments on board an aerial robot.

\begin{figure}[b!]
\centering
    \includegraphics[width=0.7\linewidth]{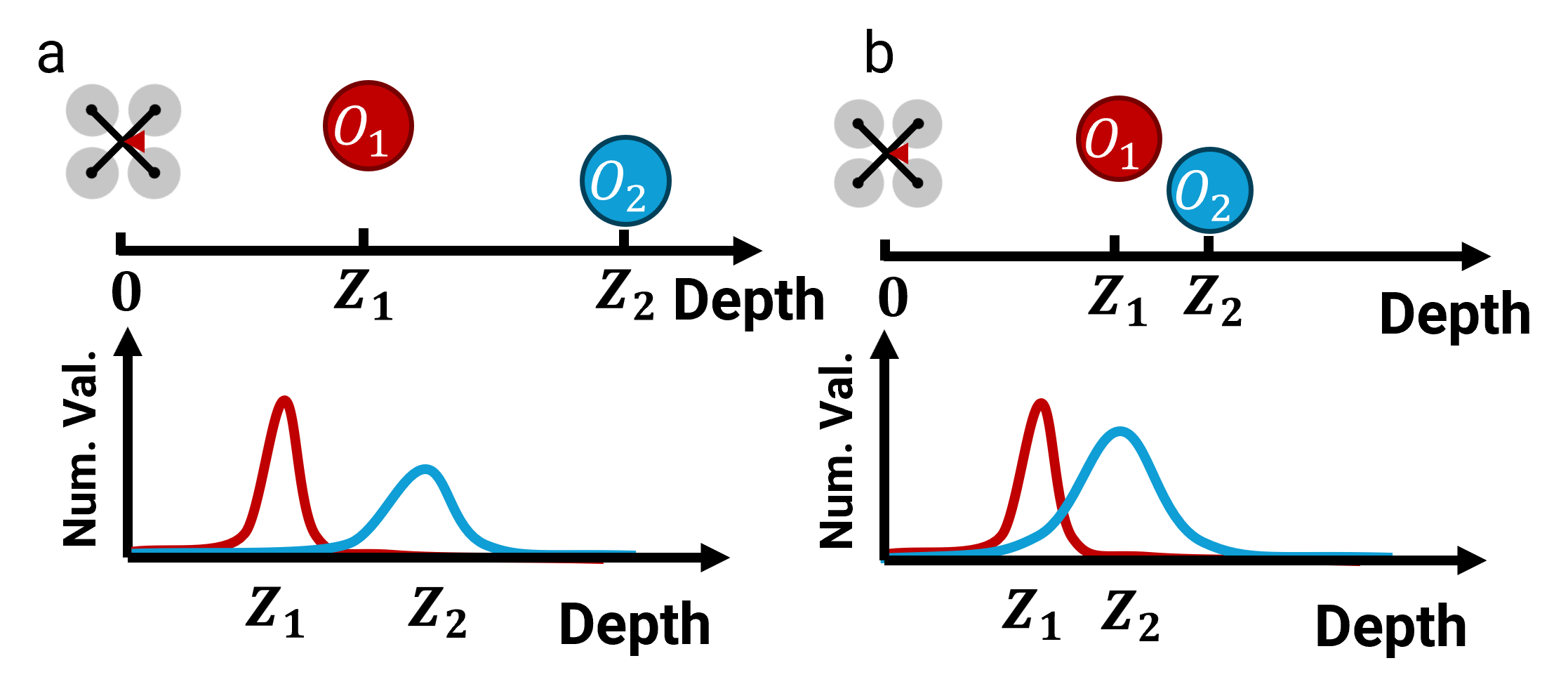}
    \caption{Robot is present in a scene (at $Z=0$) with two obstacles $O_1$ and $O_2$ at distances $Z_1$ (\textcolor[RGB]{192,0,0}{red}) and $Z_2$ (\textcolor[RGB]{15, 158, 213}{blue}) respectively. As the obstacles get closer  (a) to (b), the distance between the two modes (each obstacle is represented by a mode in the depth distribution) gets smaller, making the depth segmentation harder. This is analogous to inter-class distribution in computer vision. See Suppl. \S \textcolor{red}{S.III.} for more information.}
    \label{fig:depth_modes}
\end{figure}

\textcolor{black}{To understand this concept more formally, let us imagine that our scene consists of several obstacles, each at a distinct depth. The depth distribution of the scene can then be viewed as a multi-modal univariate distribution with each mode representing one obstacle\cite{Sanket_2018}. Effective depth segmentation involves assigning pixels to the individual modes of this distribution (segmentation), each representing a different obstacle. Without loss of generality, let us focus on the first two closest obstacles (Fig. \ref{fig:depth_modes}), which we will term as the foreground (closest) and background (all obstacles farther away). The larger the distance between foreground and background modes, the better the segmentation performance. The closer the two modes, the closer the obstacles are in real-world and the harder the segmentation problem.} While large learning models leverage enormous amounts of data to generalize, they often rely inherently on ``recognizing known-sized objects'' to obtain scale and hence fail to generalize to scenes that mimic post-disaster due to a lack of known objects\cite{ajna}.


\textcolor{black}{Prior works have inferred depth by altering the input representation, most commonly by replacing the raw image with optical flow or flow-uncertainty fields that encode motion-dependent depth cues. Broadly, such approaches can be viewed as performing foreground-background segmentation on a function of depth $\Phi(Z)$, where a careful choice of $\Phi$ can improve separability between depth modes with reduced computational cost. We will refer to this function $\Phi$ as depth cues (Suppl. Fig. \textcolor{red}{S2}).} In optical flow, $\Phi$ is coupled to motion and $Z$ can only be recovered to a scale if metric velocities are known. In optical flow uncertainty, $\Phi$ is also coupled to motion and only depth ordinality information (closer or farther without distance associations) can be recovered. Our goal in this work is to carefully craft $\Phi$ to efficiently recover metric $Z$.  

\begin{figure}
    \centering
    \includegraphics[width=\linewidth]{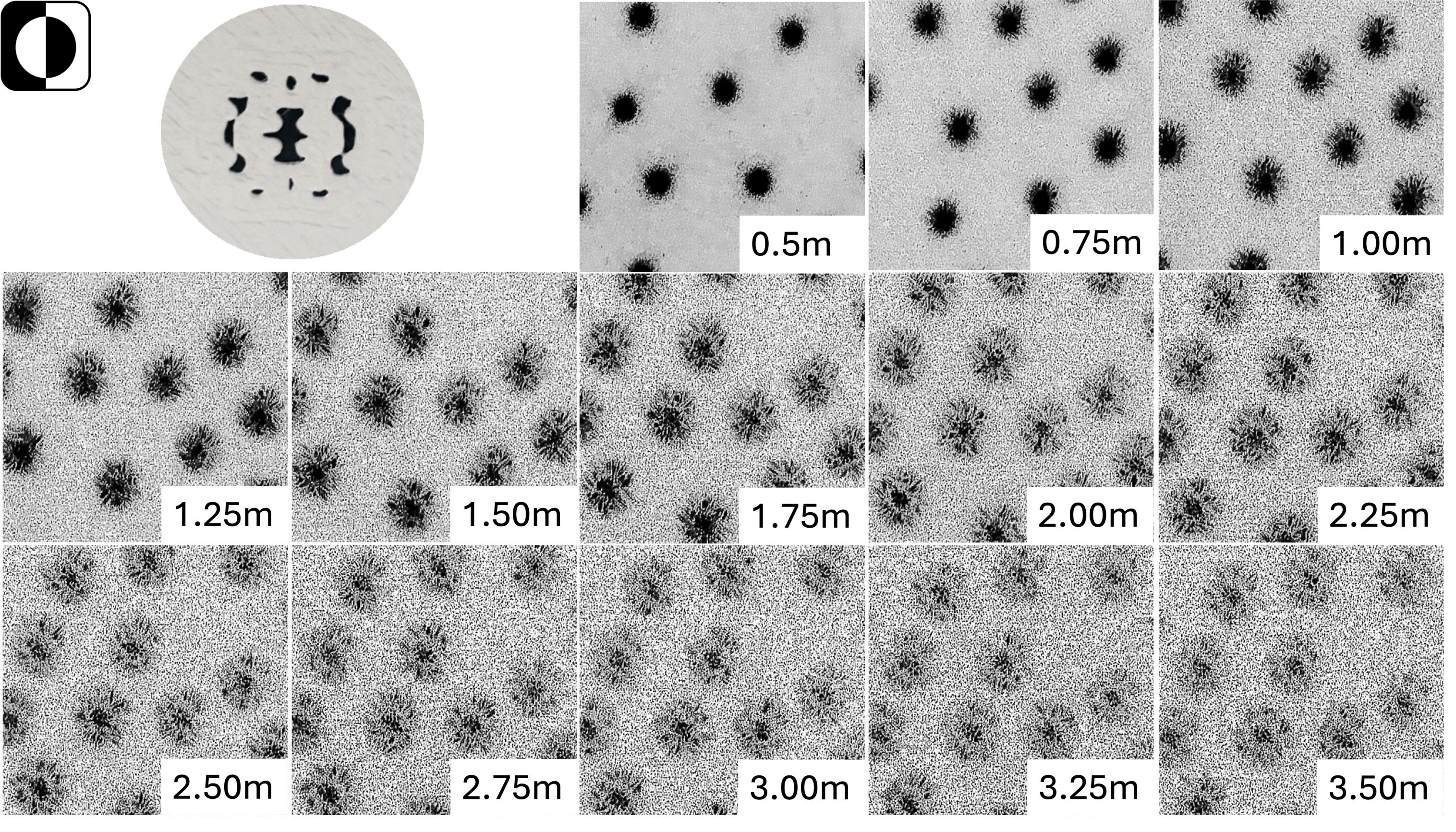}
\caption{Coded Aperture and PSF Bank, with the camera focused at 0.5m. \textit{The intensities in all the images are inverted.
\includegraphics[height=1em]{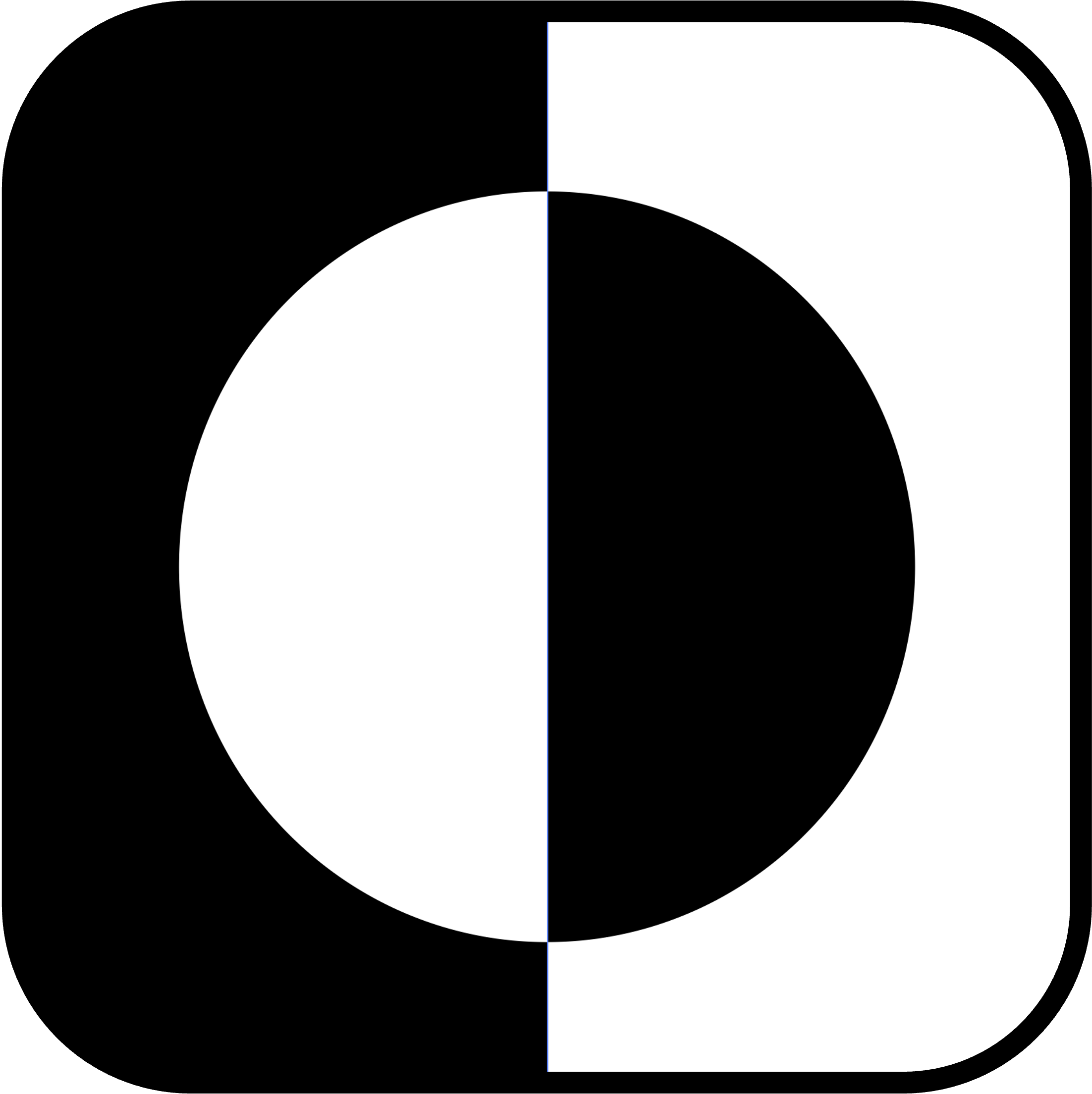}  denotes image inversion.}}
    \label{fig:psf_images}
    \vspace{5pt}
\end{figure}

\begin{figure}
    \centering
    \includegraphics[width=\linewidth]{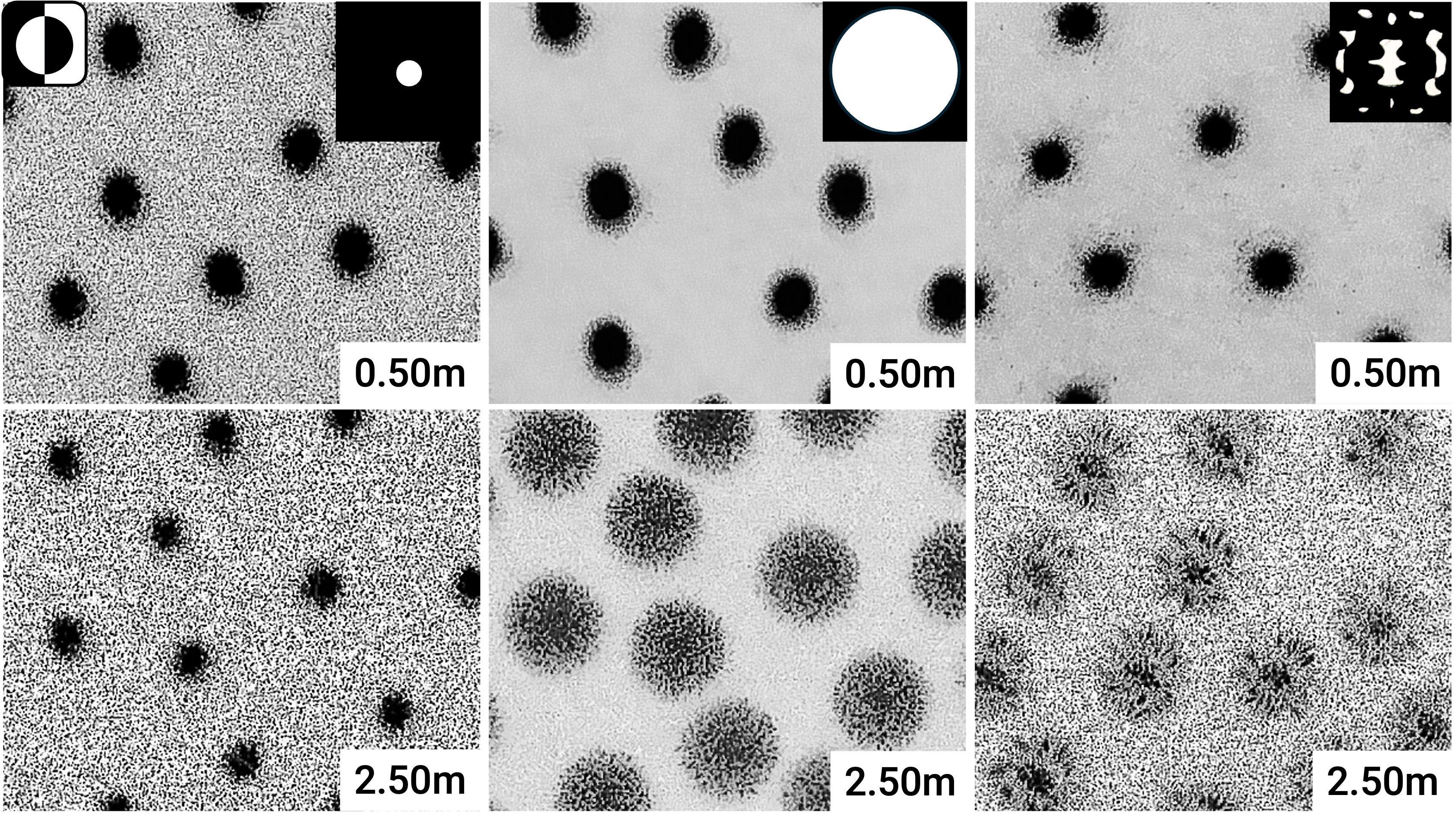}
    \caption{Left to Right: Comparison between pinhole, fully open, and coded aperture images of dot patterns observed at 0.5m and 2.5m. The pinhole aperture maintains sharpness across depths, while the fully open and coded apertures introduce depth-dependent blur. The coded aperture exhibits more pronounced variations, thereby providing stronger cues for depth estimation.}
    \label{fig:pinhole_vs_fullyopen_vs_coded}
\end{figure}

The core principle of our work relies on the fact that out-of-focus objects generate blur whose size depends on the distance from the focal plane. For a point source at depth $Z$, when the camera is focused at $Z_{f}$, blur circle diameter is
\begin{equation}
    s = \frac{f^2 \lVert Z - Z_{f}\rVert_1}{N (Z_{f} - f)Z}
\label{eq:blur_circle}
\end{equation}
where $f$ is the focal length, $N$ is the aperture number.
There are two key ways one can perform foreground-background segmentation for navigation: (a) Focus at the distance the robot needs to dodge and look for ``sharp'' regions when $N$ is low, (b) Same as (a) with the addition of explicit depth computation. In complete darkness, neither of the methods works as the data is extremely noisy to decipher focus information from it. To this end, we utilize a structured light source that projects point sources of light in a random pattern. These point sources of light can be approximated with a dirac delta function $\delta(\mathbf{X})$, where $\mathbf{X}$ represents the 3D location of the point projected on the world. Theoretically, by imaging the scene illuminated by a structured light source is enough to recover depth based on the sizes of $s$ per point, as shown in Eq. \ref{eq:blur_circle}. However, in practice, commonly used simple circular apertures do not render mathematically stable conditions for accurate depth estimation\cite{coded_aperture_depth_mit}, since the blur circle size can change with image artifacts and imperfections in the structured light/aperture. To address this, we utilize a coded aperture from \cite{coded_aperture_depth_mit} to produce depth dependent distinctive spatial patterns in the defocus blur. Unlike a uniform blur disk, these patterns vary more significantly with depth and are easier to disambiguate during estimation. Fig.~\ref{fig:psf_images} shows the coded aperture used in our setup, and images of dot patterns projected from different depths onto a plane wall. Fig.~\ref{fig:pinhole_vs_fullyopen_vs_coded} compares outputs from a pinhole aperture, a fully open circular aperture, and the coded aperture, for a planar wall illuminated by the structured light source. Coded apertures substantially improve depth estimation accuracy even when used with small-format sensors, making them suitable for aerial robots navigating in darkness.

Finally, to obtain metric depth from coded structured light images, we utilize a deep convolutional neural network. Our choice of using a neural network is for better robustness and speed compared to classical methods\cite{coded_aperture_depth_mit}. The key challenge lies in training the network to associate specific defocus characteristics with their corresponding depths without requiring extensive real-world labeled data. 

With this goal, we present a novel data generation strategy in simulation that enables zero-shot generalization of the model to real-world environments. By carefully modeling the optical characteristics of our coded aperture system and synthesizing realistic training data, we achieve robust depth estimation performance without the need for ground-truth depth collection across various scenes.


\subsection{Training Data Generation}
\label{sec:data_gen}
To train a learning-based depth estimation model, we construct a synthetic dataset that captures the defocus characteristics of our coded aperture system. The key idea is to model the Point Spread Function (PSF) at different depths and use these PSFs to synthesize training images with ground truth depth in simulation.

\begin{figure}
    \centering
    \includegraphics[width=0.9\linewidth]{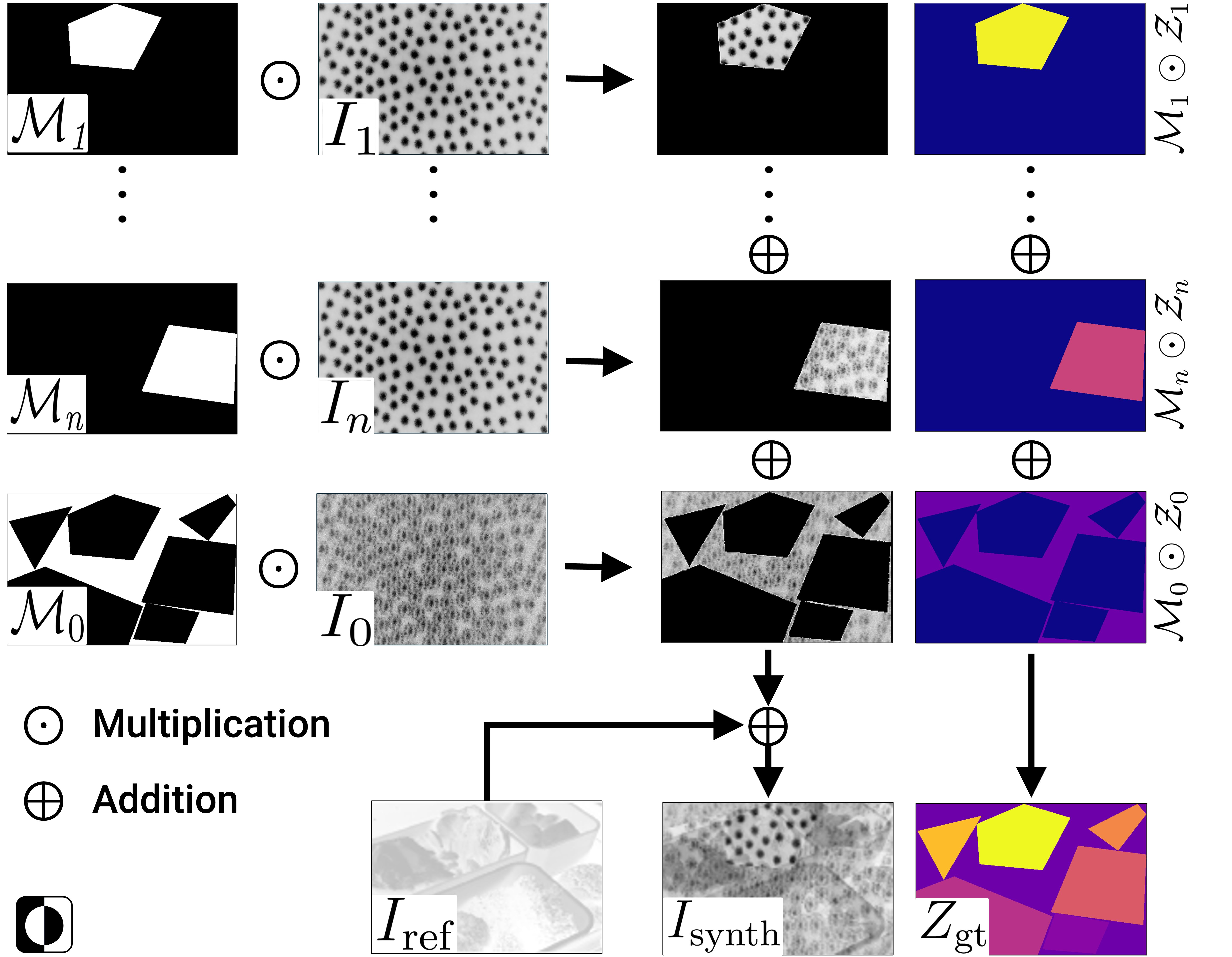}
    \caption{Synthetic data generation pipeline. $I_i, I_\text{ref}, I_\text{synth}$ are inverted.}
    \label{fig:datagen}
\end{figure}

We begin by projecting a dot pattern onto a planar wall and capturing images at multiple known depths which we will refer to as the \textit{depth planes}. We observed that using real-images worked better for cheaper lenses used in robotics due to large amounts of distortion and low-quality diffractive correction that are hard to model in simulation. For higher quality lenses, mathematical models without any real calibration data would be sufficient. Each captured image corresponds to the defocus response of the optical system at a particular depth, thereby forming a PSF bank (Fig.~\ref{fig:psf_images}). To synthesize a large amount of data, we create depth images with randomized polygonal obstacles at various depths. Note that no ray tracing is used for these image creations; it is a simple random image generation. We utilize the corresponding images (corresponding to the depth of the polygon) from the PSF bank using alpha matting against a randomly generated dark and noisy background. Fig. \ref{fig:datagen} shows our synthetic data generation pipeline. The complete process is explained in detail next.



\textcolor{black}{Let the images of the wall captured at distances $Z_1, Z_2, \ldots, Z_n$ (which we refer to as discretized depth planes) be denoted by $I_{1}, I_2, I_3, \ldots, I_n$. These images obtained from the camera can be modeled as}
\begin{equation}
    I_{i} = I_{o} \circledast h(Z_{i})
\end{equation}
where $I_o$ is the All-In-Focus (AIF) image of the same wall and \textcolor{black}{ $\circledast$ is convolution operator}. Images from different distances have a different amount of blur in the image, hence changing the shape of the dot pattern in the image based on depth of the dot. This set of images is called the \textit{calibration set} denoted as $I_i$.  

For creation of the training data, we take reference images from the MS-COCO dataset \cite{lin2015microsoftcococommonobjects} $I_{\text{ref}}$ to serve as the background. The background intensity is scaled down to mimic low-light scene and gaussian noise is added to simulate noise in captured image in low-light. We then overlay random crops from different calibration images onto this reference image \textcolor{black}{to represent obstacles}. To do this, we generate a random depth map $Z$ (from the \textit{depth plane} ranges) with polygonal obstacles. The polygonal obstacles vary in location, size and number of vertices. Let $\mathcal{M}_i$ be a mask which is $1$ for all the pixels with depth $Z_i$ and $0$ elsewhere. Note that $\mathcal{M}_i$ can have zero or $m$ polygons. $\mathcal{M}_0 = \left(\neg \left(\cup_{i=1}^n \mathcal{M}_i\right)\right)$ represents the region that are not occupied by polygons which we designate as the ``background'' region.
To obtain the final image, we mask the calibration image at $Z_i$ with $\mathcal{M}_i$ and alpha blend it with the background $I_{\text{ref}}$. 
The synthetic training image $I_{\text{synth}}$ is then constructed as:
\begin{equation}
I_{\text{synth}} = w_{ref}I_{\text{ref}} + \sum_{i=1}^{n} \alpha_i \cdot \mathcal{M}_i \odot I_i + \alpha_{0}\cdot \mathcal{M}_0\odot I_{0}
+ \mathcal{N}(0, \sigma^2)
\end{equation}







Here, $w_{ref}$ represents the weight factor for scaling the intensities in the reference image, which accounts for the camera sensitivity and ambient global illumination. $\alpha_i$ is the blending weight for the $i^{\text{th}}$ depth plane; this parameter represents the material reflectivity characteristics. The zeroth index for $I$, $\alpha$ and $\mathcal{M}$ refers to all the pixels in the image not occupied by polygons (so-called ``background'', loosely speaking, since this value can be closer than the obstacles as well to represent a gap-like scenario seen in \cite{Sanket_2018}).  $\mathcal{N}(0, \sigma^2)$ represents pixelwise Gaussian noise drawn from a normal distribution with mean 0 and variance $\sigma^2$. The corresponding ground truth depth map $Z_{\text{gt}}$ is computed as: 
\begin{equation}
Z_{\text{gt}} = \sum_{i=1}^{n} \mathcal{M}_i \odot \mathcal{Z}_i + \mathcal{M}_0\odot \mathcal{Z}_{0}
\end{equation}
Here, $\mathcal{Z}_i$ is an array of the size of the image $I_{\text{synth}}$ filled with $Z_i$ depth values, similarly $\mathcal{Z}_{0}$ represents the ``background'' (non-polygonal area) of depth $Z_0$. Our approach generates synthetic training pairs $(I_{\text{synth}}, Z_{\text{gt}})$ where each input image 
contains multiple regions with different defocus blur patterns, corresponding to various randomly generated polygons. The model can thus learn to associate the various blur characteristics with their corresponding depths across the diverse training samples.

By repeating this procedure multiple times, we generate a dataset of 100K images, which serve as the training data for our depth estimation network, \textit{AsterNet}. Importantly, our model is never trained on real-world scenes, yet it generalizes effectively to such environments during inference due to enough domain randomization. This demonstrates that the coded aperture defocus patterns, combined with active lighting, provide a sufficiently rich cue for learning depth in challenging dark settings. 


\subsection{\textit{AsterNet} Network Architecture and Training}
\label{subsec:model_training}





Our \textit{AsterNet} model is designed as an encoder--decoder architecture with skip connections from encoder to decoder, following the style of U-Net~\cite{ronneberger2015unetconvolutionalnetworksbiomedical}. The encoder is based on DenseNet121~\cite{huang2018denselyconnectedconvolutionalnetworks}, where the first convolution layer is modified to accept a single-channel input, while all subsequent layers remain identical to the original DenseNet121 design (see Suppl. Fig. \textcolor{red}{S1} and Suppl. \S \textcolor{red}{S.I}). 

The decoder consists of a series of upsampling blocks, each performing bilinear upsampling with a scale factor of 2, followed by Batch Normalization (BN) and a ReLU activation. A $3 \times 3$ convolution with stride $S=1$ and padding $P=1$ is then applied, after which another BN-ReLU pair is used. Skip connections are taken from the output of each dense block from the encoder and resized before augmenting with decoder layers. Furthermore, to enhance generalizability to the real-world, we train our network with a self-supervised uncertainty loss from \cite{ajna}. The final loss function is:

\begin{equation}
\mathcal{L} = \text{Huber}_\delta(\tilde{Z}, Z_{\text{gt}}) \cdot e^{-\log \Upsilon^2} + \frac{1}{2} \log \Upsilon^2
\label{eq:heteroscedastic_loss}
\end{equation}

Here, $\tilde{Z}$, $Z_{\text{gt}}$ are the predicted and ground truth respectively, and $\Upsilon$ is the predicted variance. The model is trained on 100K synthesized images (See $\S$\ref{sec:data_gen}) using the loss function from Eq. \ref{eq:heteroscedastic_loss}, with an ADAM optimizer, with an initial learning rate of \textbf{$10^{-3}$} and, a batch size of \textbf{$32$}. The network was trained for 50 epochs with a learning rate decaying by a factor of 0.5 every 10 epochs.

\section{Parsimonious Approach For Navigation Using Depth Cues}
\label{sec:parsimonious}
\textcolor{black}{If the robot is tiny, running a deep learning-based depth model might not be feasible. For the navigation task, the parsimonious solution (a solution that uses the minimum amount of information and computation to achieve the task) is to distinguish (segment) foreground obstacles that need to be dodged from the background that can be ignored. To this end, our approach starts by setting focus to $Z_{\mathcal{F}}$ (or Foreground distance), the distance at which the obstacles need to be dodged for a maximum robot velocity (we set this to 0.5m-1m depending on the scene complexity and desired robot speed). 
For foreground-background segmentation (obstacle detection), we can detect the focus points (corners in computer vision) of known size from our structured light. To do this, we utilize two standard computer vision methods \cite{szeliski2022computer}: (a) Difference of Gaussian ($DoG$) method and (b) Template Matching ($TM$) method. In $DoG$, we subtract Gaussian Blurred versions of the image at two different scales ($\sigma$ which is set by desired blur circle size we want to detect), similar to SIFT feature detector\cite{lowe2004distinctive}. The higher the value, the higher the response, implying a high likelihood of an obstacle at the particular distance. Here, we are detecting blur circles of different sizes using simple $DoG$ operators, which in-turn give us depth. If we need more fine-grained control, a more expensive option is $TM$, where an image template of the focused point (or even the defocused point by focusing on the background) is matched using Template Matching with Phase Correlation similar to stereo matching\cite{szeliski2022computer} to match with the closest depth-based blur pattern based on a calibration sequence (similar to that in \S \ref{sec:asterNet}). This method ``looks'' for the closest pattern from the calibration bank by brute-force matching to obtain depth.}

\section{Navigation Policy}
Our navigation approach is inspired from Ajna \cite{ajna}, wherein we threshold the depth map to dodge closer obstacles, while treating the farther ones as background using a simple potential field planner. Similar to Ajna, the objective of the robot is to keep going straight, with lateral or vertical movement to only dodge obstacles in the way. We segment the depth map into foreground and background by thresholding. For the parsimonious solutions presented in \S \ref{sec:parsimonious}, we utilize the segmented map to identify the background region (free space) and align to it.

\section{Experimental Results And Discussion}
\label{sec:expts}
We divide our experiments into real-world benchtop experiments and real-world flight experiments. In the first set of experiments, we validate the effectiveness of our approach and compare it with other state-of-the-art depth estimation methods. In the second set, we validate our depth estimation method as a tool for quadrotor navigation in complete darkness. \textit{It is pivotal to note that our approach is trained completely in simulation without real-world data from any of the scenes it was tested/deployed, leading to a complete sim2real and zero-shot generalization evaluation.}

\subsection{Hardware Setup}
Our sensing setup uses an Arducam 477P High Quality Camera Module with the Sony IMX477 $1/2.3"$ sensor mated to a 16mm $f/1.4$ lens. We keep the aperture wide open at  $f/1.4$ unless otherwise specified. The lens's back-element is covered with a 3D printed black coded aperture mask from \cite{coded_aperture_depth_mit}. The coded aperture is 0.3mm thick and has a diameter of 6mm. We use a Bambu Labs X1 Carbon with black eSUN PLA+ 1.75mm 3D Filament to fabricate the coded aperture. All the ground truth measurements of depth are obtained with an Intel$^\text{\textregistered}$ Realsense$^\text{TM}$ D435i. We utilize the structured light projector from an  Intel$^\text{\textregistered}$ Realsense$^\text{TM}$ D435i module with cameras blacked off with electrical tape for a cost-effective and easy to procure solution, but this can be replaced with any off-the-shelf module.  

The robot used in the experiments is a custom-built platform called PeARCorgi210 (Fig. \ref{fig:Banner}) that has a 210$mm$ diagonal wheelbase. All the lower-level control algorithms run on the Holybro Pixhawk 32 V6 Flight Controller using ArduPilot Copter v4.5.1. The ArkFlow optical flow sensor is used to enable hover. We use a small LED on the bottom for enabling optical flow operation, this does not alter the scene brightness for our obstacle avoidance sensors or stack. All the higher-level perception, decision making and control commands are computed on the onboard NVIDIA Jetson Orin$^\text{TM}$ Nano running Ubuntu 20.04.6 LTS and are sent to the flight controller using MAVLink. The perception stack takes input from the onboard aforementioned camera.


\subsection{Environmental Setup}
\begin{figure*}
    \centering
    \includegraphics[width=\linewidth]{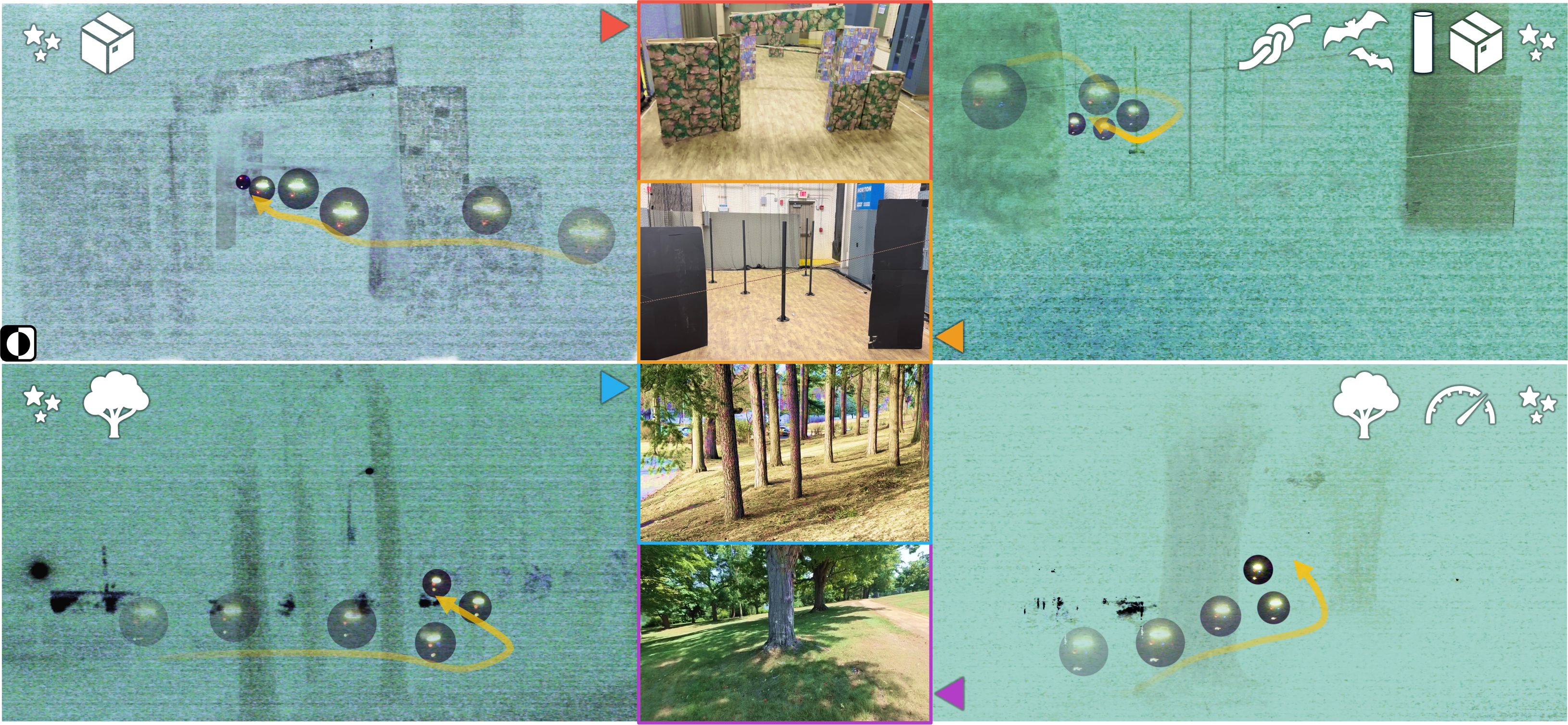}
    \caption{Sequence of images of a quadrotor navigating through different environments, with obstacle types denoted by symbols shown in the top left of each subfigure. Quadrotor and arrow transparency shows progression in time. The images are inverted for better clarity, images are taken at f/2.8, 1/100s and ISO 102400 on a Nikon D850 DSLR. The scene in normal lighting is shown in the center column. (See Table  \ref{tab:exp}).}
    \label{fig:hardwareruns}
\end{figure*}

The flight experiments were tested in a combination of indoor and outdoor settings  (Table \ref{tab:exp} and Fig. \ref{fig:hardwareruns}). We used forests with varying tree densities, tree dimensions and textures to test the robustness of our approach in outdoor scenes. Our outdoor scenes denoted by \texttt{Forest$_1$} (Fig. \ref{fig:hardwareruns} bottom right) and \texttt{Forest$_2$} (Fig. \ref{fig:hardwareruns} bottom left), have traversabilities \cite{traversability} $18.30$ and $8.20$.  Indoor flight tests were performed in a netted facility of dimensions $11 \times 4.5 \times 3.65m$. We construct two different scenarios with a variety of obstacles (Fig. \ref{fig:hardwareruns}). In the first experiment (we call \texttt{Boxes}, Fig. \ref{fig:hardwareruns} top left), the obstacles are made of cuboidal cardboard boxes of sizes ranging from $1.15 - 1.28m$ with rock and moss textures stuck on them \cite{edgeflownet}, and arranged in a manner such that the resulting traversability is $4.80$. In the second setup, termed \texttt{Dark Objects} (Fig. \ref{fig:hardwareruns} top right), the same cuboidal obstacles were covered with matte black stickers to create extremely low-reflectivity conditions. This scenario also included PVC pipes (height: 1.56$m$ , \diameter: 0.043$m$) wrapped in the same black material, along with a thin rope (\diameter$6.35mm$) that was predominantly black with small red and yellow patches. It is important to note that all our indoor experiments were conducted with all lights OFF, the overall intensity was less than 1 milli-lux (starlight on a new moon night). We had our motion capture system turned OFF in all the experiments since this would increase our ambient illumination from IR lights. For outdoor experiments, the closest street lights (if any) were covered with a black cloth and the overall illumination levels were about 0.001-0.1 lux.

\subsection{Real-World Flight Tests}
\label{sec:hardwarerun}

\begin{table}
\vspace{3mm}
\caption{Quantitative evaluation of \textit{AsterNav} in real-world flight experiments. (See Fig. \ref{fig:hardwareruns}).}
\centering
\resizebox{\columnwidth}{!}{%
\begin{tabular}{lcccc}
\toprule
Scene        & Traversability & Brightness (lux) & Temperature ($^{\circ}$C) & SR \\
\midrule
\texttt{Forest$_{1}$}   & 18.30 & 0.1 & 13.9 & 5/5 \\
\texttt{Forest$_{2}$}       & 8.20 & 0.1   & 16.1 & 15/15 \\
\texttt{Boxes}        & 4.80 & 0.001    & 22.2 & 14/15 \\
\texttt{Dark Objects} & 3.50 & 0.001    & 21.7 & 14/15 \\
\bottomrule
\end{tabular}}
\label{tab:exp}
\end{table}

For testing the robustness of our approach in the real world, we tested it in both indoor and outdoor scenarios, with results shown in Table \ref{tab:exp}. We use the Success Rate ($SR$) metric to analyse the performance of our perception and navigation approach. $SR$ is the ratio of the number of times the robot navigated through the scene without hitting any obstacles to the total runs in each scene. For the \texttt{Boxes} scene, we used boxes of varying sizes as vertical obstacles, and a thin box and pole as horizontal obstacles. This was done to show that the robot can dodge both horizontally and vertically. In this scene, we achieved a success rate of 14 out of 15 runs. The \texttt{Dark Objects} scene had poles and boxes placed vertically, with the 6.35mm\diameter rope placed horizontally. In this scene, we achieved 14 successful runs out of 15 trials. For both indoor scenes, the model performed exceptionally well, with failures occurring due to erroneous tuning of potential field parameters. Outdoor scenes (\texttt{Forest$_1$} and \texttt{Forest$_2$}) were less cluttered compared to the indoor setups, and the robot successfully avoided obstacles in all trials. In \texttt{Forest$_1$}, we conducted the runs at a higher speed (peak 3.5 \( ms^{-1} \)), to demonstrate the capability of our compute-efficient approach to operate at fast flight speeds. While the system could likely sustain even higher velocities, we refrained from pushing further due to safety considerations. At higher speeds, the available response time for manual intervention decreases significantly, making it difficult for the operator to halt the robot in case of an imminent collision. This challenge was further exacerbated by the fact that the scene was completely dark, which made visually tracking the robot and obstacles to issue timely manual overrides even harder. For the same reason, we limited the number of trials at this speed to 5.  Overall, the approach achieved a success rate of $43/45$ ($95.5\%$). The high-speed runs from \texttt{Forest$_1$} are excluded from this aggregate, due to the statistically small sample size. 

\subsection{Benchtop Experiments for Depth Estimation Performance}



\begin{figure}[t!]
    \centering
    \includegraphics[width=\linewidth]{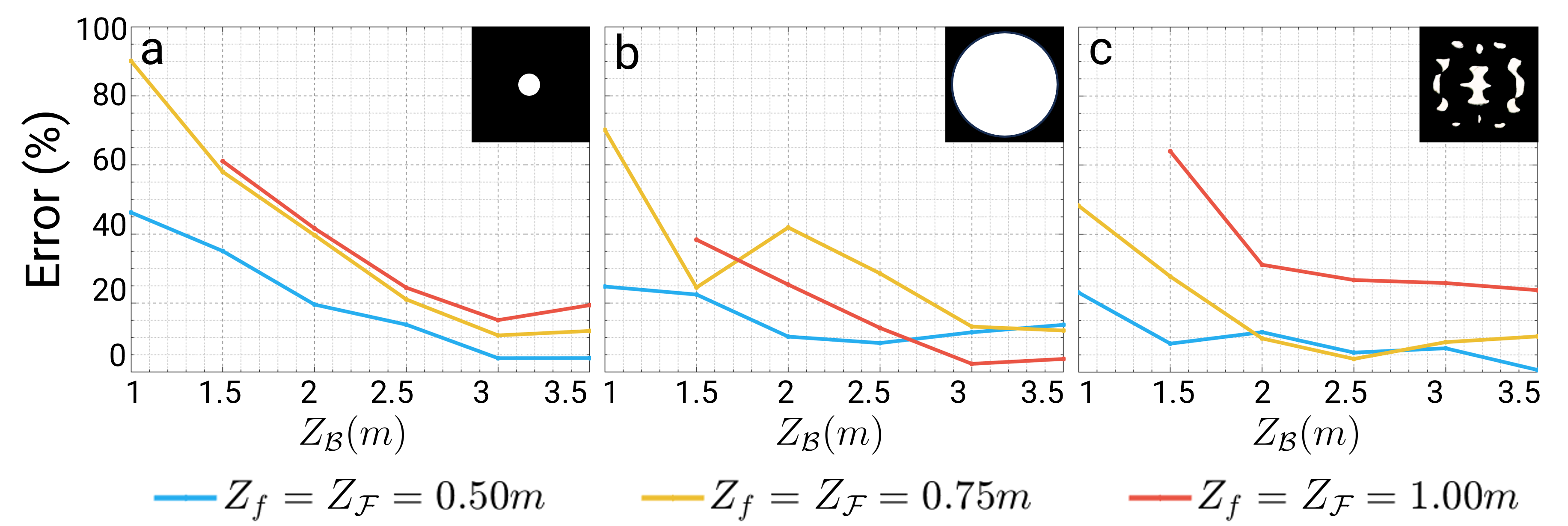}
    \caption{Variation of $l_1$ error with background distance $Z_\mathcal{B}$ for different foreground distances $Z_\mathcal{F}$ for (a) Pinhole ($f/8$),  (b) Fully Open ($f/1.4$), and (c) Coded Aperture. Focus is fixed at the foreground ($Z_f = Z_\mathcal{F}$).}
    \label{fig:ca_fo_ph}
\end{figure}

We obtain images of a foreground obstacle at $Z_{\mathcal{F}}$ where it is focused (thus $Z_{f} = Z_{\mathcal{F}}$) while the background is at $Z_{\mathcal{B}}$. We use the same textures from the \texttt{Boxes} experiment for foreground and background. The distance between $Z_{\mathcal{F}}$ and $Z_{\mathcal{B}}$ determines the scene complexity (denoted by a traversibility score\cite{traversability}). Intuitively, the smaller the distance, the harder it is for the robot to navigate. We ensure that the foreground covers roughly 50\% of the pixels in the image to minimize bias in our experiments. We compare our accuracy results as $l_1$ error with respect to ground-truth (obtained manually). We evaluate the performance of our system and compare it with fully-open aperture ($f/1.4$) and pinhole apertures ($f/8$) with the same camera and lens combination (Fig. \ref{fig:ca_fo_ph}). Our network was re-trained for every $Z_f$ (focus distance) for each of the case since the network learns to decipher depth based on blur patterns from the PSF, which change drastically by changing the focus distance. The error percentage in prediction decreases as the distance between foreground and background increases. \textcolor{black}{As visible in Fig.\ref{fig:ca_fo_ph}, the error percentage is lower in coded aperture setup for $Z_f=0.5m$ and $Z_f=0.75m$, as compared to fully open and pinhole apertures. Since the depth of field is much larger for $Z_f = 1m$, the error is high for all the apertures, mainly limited by camera sensor size.}
Our coded aperture system gives the best segmentation results for $Z_f=0.5m$, which is why we chose the same $Z_f$ for our hardware runs in the \S \ref{sec:hardwarerun}. 
\textcolor{black}{In most scenarios ($Z_f=0.5m$ and $Z_f=0.75m$), the coded aperture provides higher accuracy than the pinhole and open aperture due to strong depth cues that inform \textit{AsterNet} model.
Interestingly, the performance of fully open aperture for $Z_f=1m$ is better than both pinhole and coded aperture since it lets the most light into the sensor. This enables weak reflections from longer range to be detected more effectively than pinhole or coded apertures, both of which let in less light (Fig. \ref{fig:pinhole_vs_fullyopen_vs_coded}).}



\begin{figure}[t!]
    \centering
    \includegraphics[width=0.6\linewidth]{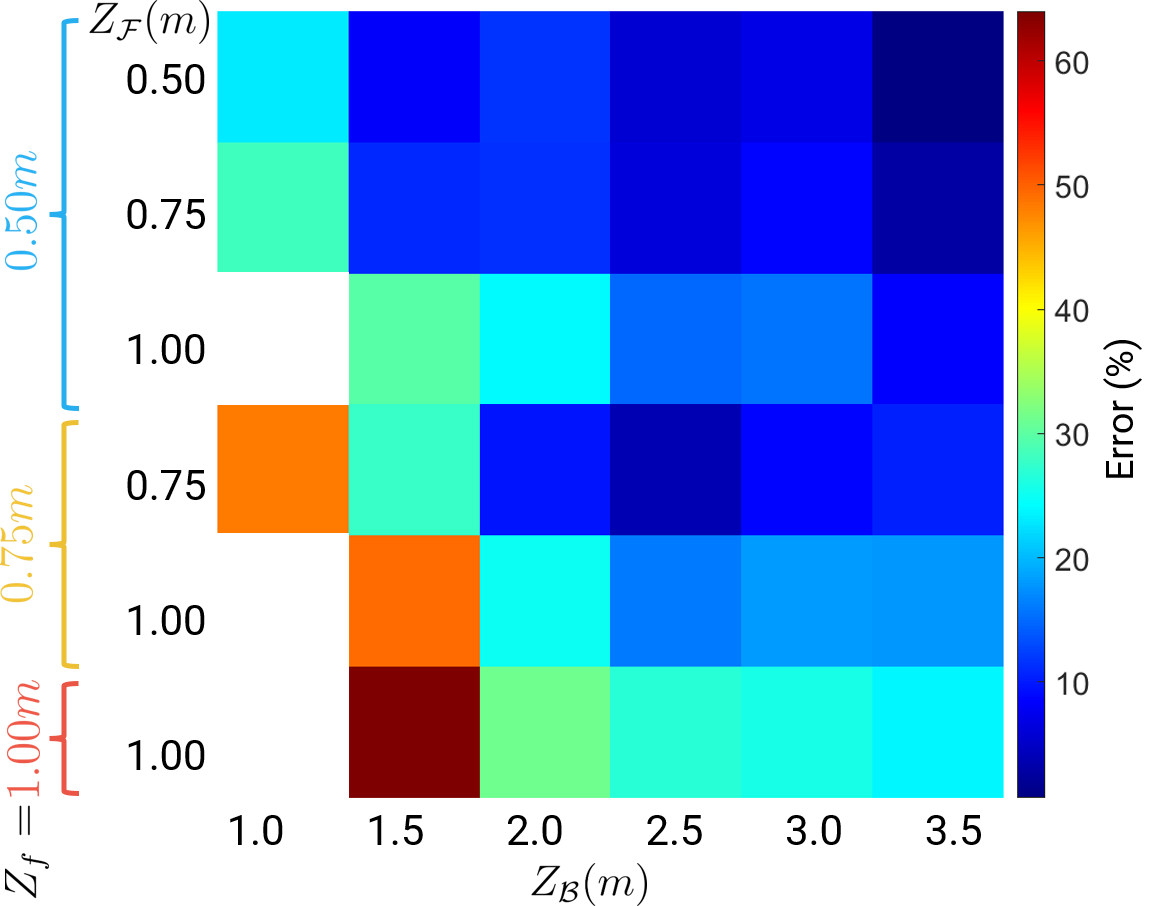}
    \caption{Coded Aperture $l_1$ error variation with $Z_\mathcal{F}, Z_\mathcal{B}, Z_f$.}
    \label{fig:zfzbvariation}
\end{figure}

Furthermore, we evaluate performance with variations in $Z_\mathcal{F}, Z_\mathcal{B}, Z_f$ (Fig. \ref{fig:zfzbvariation}). Obviously, as $Z_\mathcal{B}\uparrow$ or $Z_\mathcal{F}\downarrow$, the performance increases since the foreground-background disparity increases. With $Z_\mathcal{F}\uparrow$ or $Z_f\downarrow$, the opposite effect occurs making our method perform less optimally. For best results, a myopic vision is preferred which is applicable to smaller and more agile robots that require a smaller sensing range\cite{edgeflownet}.

\begin{figure}[t!]
    \centering
    \includegraphics[width=\linewidth]{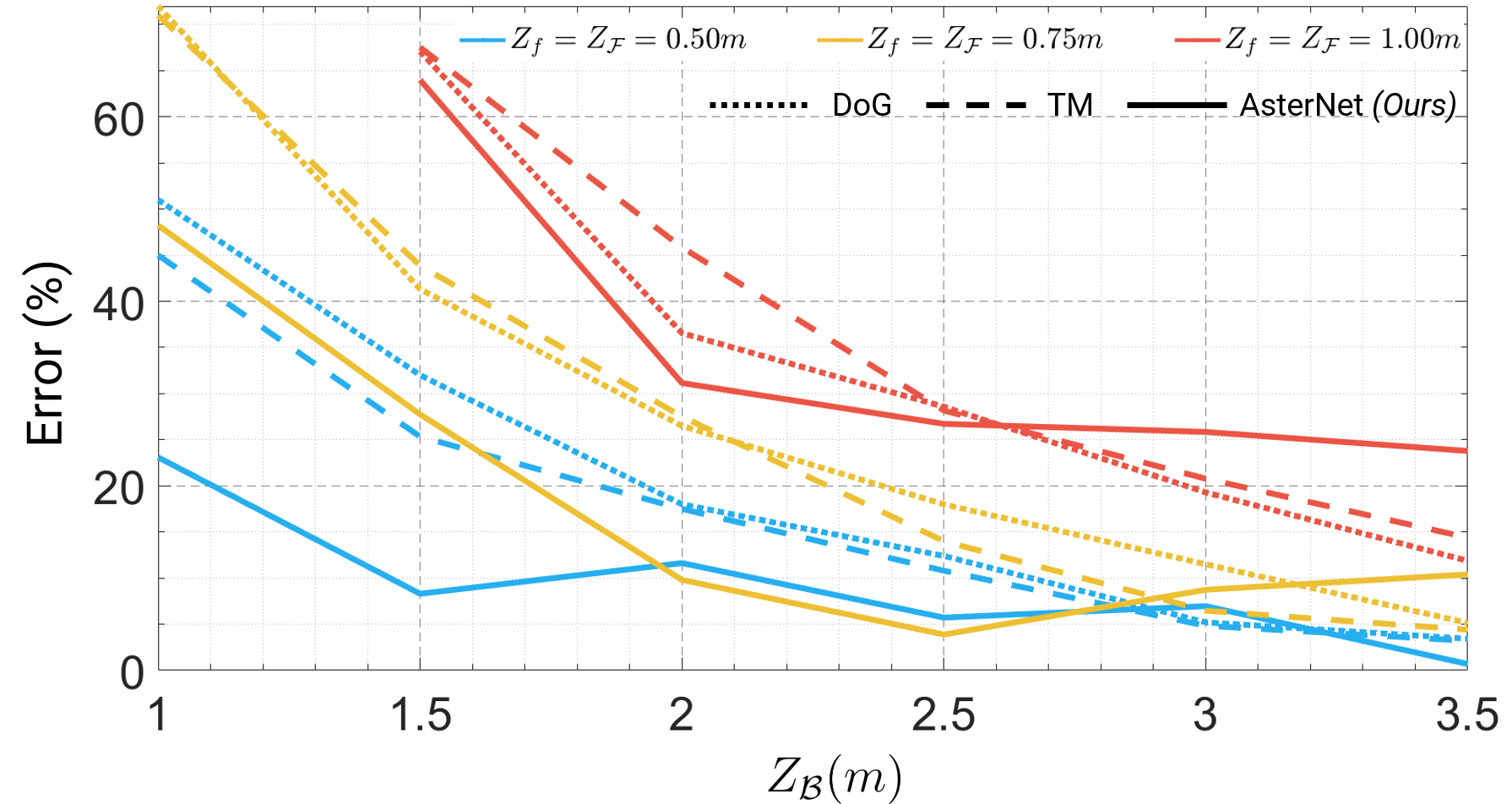}
    \caption{\% $l_1$ error for $DoG$, Template Matching ($TM$) and \textit{AsterNet (Ours)}.}
    \label{fig:dogtm}
\end{figure}

We also compare our approach with classical Difference of Gaussian ($DoG$) and Template Matching ($TM$) based segmentation methods (Fig. \ref{fig:dogtm}). Although our method predicts a dense depth map, the $DoG$ and $TM$ methods only produce foreground-background segmentation, in these methods, we assign the known depth values to foreground and background pixels, helping us convert from segmentation map to a depth map.

\begin{figure}
    \centering
    \includegraphics[width=\linewidth]{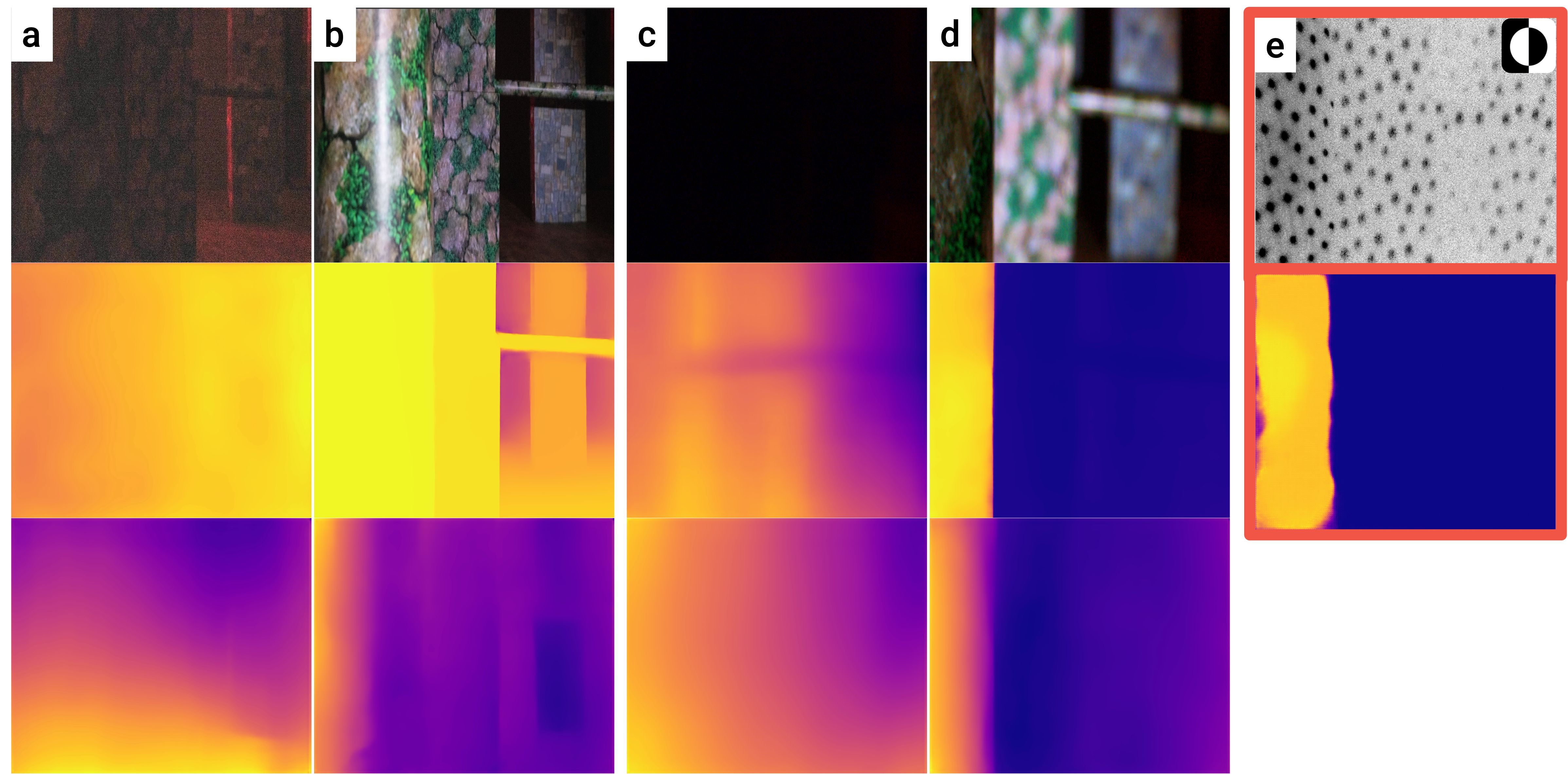}
    \caption{Qualitative comparisons of depth estimation using DepthPro \cite{depthPro} (row 2), MiDaS-Small \cite{midas} (row 3), \textit{AsterNet (Ours) (row 2, e)}.
    Left to right: (a,b) from the D850 camera without and with external illumination, (c,d) from the STARVIS camera under the same conditions, and (e) from the camera with coded aperture used in our setup (with inverted intensities).
}
    \label{fig:starvis_depth_comparison}
\end{figure}

We also qualitatively compare results (Fig. \ref{fig:starvis_depth_comparison}) of our approach with a Nikon D850 DSLR with a Tamron 15-30mm $f/2.8$ lens, the largest sensor for robotics cameras, the Arducam Sony STARVIS 2 IMX585 $1/1.2"$ camera mated to the same lens as our setup, and lastly, the same camera we used without an IR filter and with/without structured light. Since the D850 and STARVIS operate in the visible spectrum, a white LED torch providing 4 lux was used to brighten the scene. We test on images from these sensors with state-of-the-art depth models of DepthPro \cite{depthPro} and MiDas \cite{midas} (without any fine-tuning or re-training).

\begin{figure}
    \centering
    \includegraphics[width=0.88\linewidth]{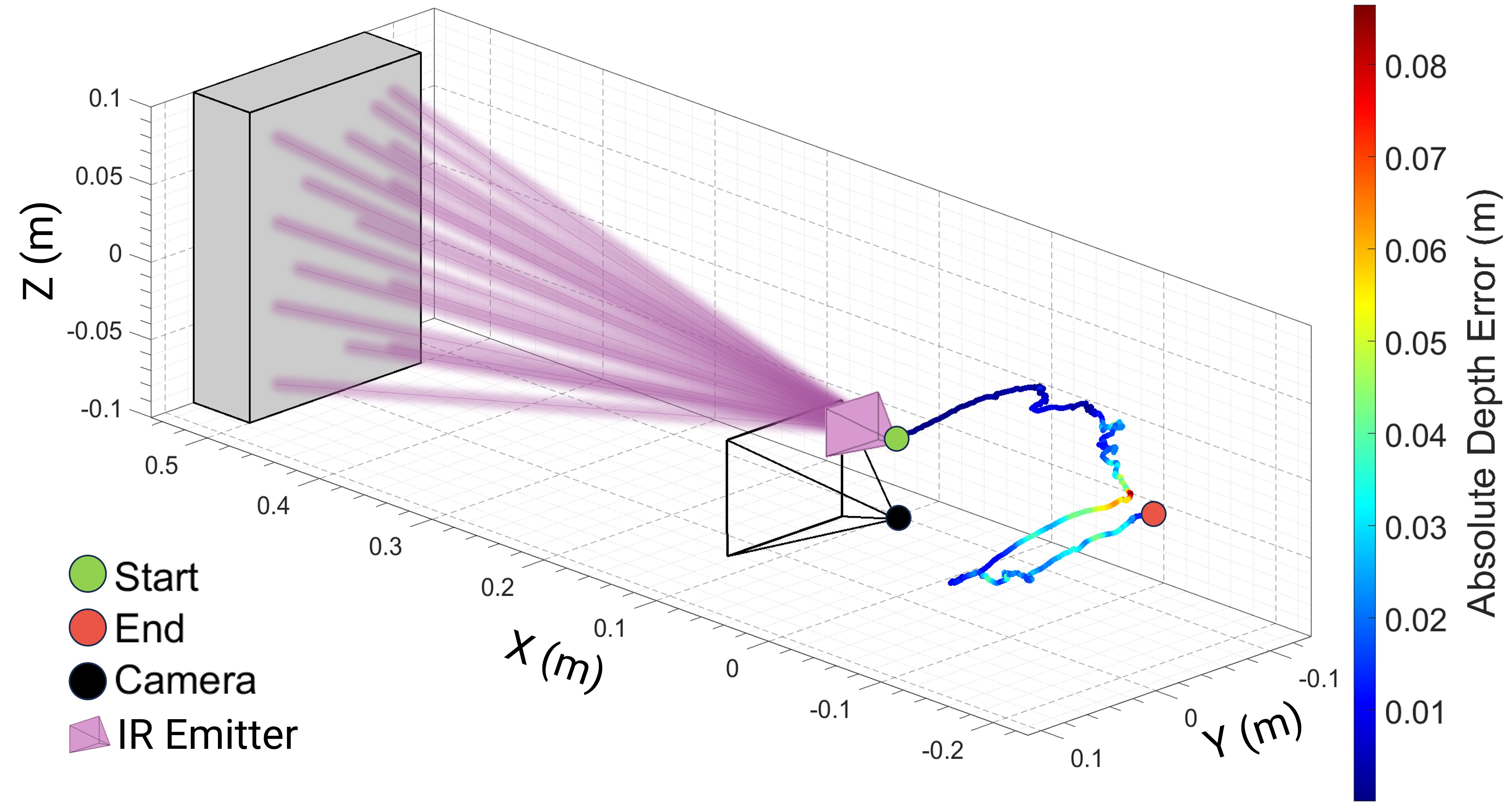}
    \caption{$l_1$ Depth Error for variation in emitter-camera relative pose simulating mis-calibration and mis-alignment showing \textit{AsterNet}'s robustness.}
    \label{fig:realsense}
\end{figure}

\subsection{Discussions and Future Work}
\label{subsec:discussion}


We observed high-depth accuracy ($l_{1}$ of 0.174$m$) up to 2$m$ range (with $Z_{f}$ as 0.5$m$), beyond which the focus blur wasn't sufficient. This error is largely attributable to the discretization of our training data at 0.25$m$ intervals, which forces the model to interpolate for objects that do not lie exactly at those depths. Importantly, the model was never trained on real-world images; the reported error is obtained entirely from real-world obstacle placement, making this a \textit{true zero-shot} evaluation. \textcolor{black}{Under this setting, our approach achieves $8.7\%$ $l_1$ error over a maximum operational range of 2$m$. Furthermore, fine-tuning our neural network on real-world data will reduce the prediction error enhancing the accuracy of the system.} Since our robot size is 210mm, the operational range is $\sim$ $14 \times$ robot size and hence the available depth is sufficient for safe navigation in cluttered environments. \textcolor{black}{We believe a task-centric co-design of coded aperture can enable a larger operational range and we see this as a promising potential for future work. Using a more powerful dot projector or a camera sensor with higher IR sensitivity will increase the depth range of our approach. Also, a higher focal length lens (or a zoomed lens) has a better Depth of Field (DoF) effect over a larger depth range. Using these methods, we can enable high-speed robot autonomy in darkness using the increased depth sensing range.} Furthermore, compared to a wide aperture, which also has depth cues and a small aperture, which lacks depth cues, our coded aperture method performs $38.6\%$ better than the fully open one and $58.2\%$ better than pinhole aperture, due to stronger physics constraints through passive computation\cite{pawar2025blurring}. We also observe that a simple parsimonious approach is about $50\%$ as good with $\sim 10ms$ computation time (on CPU) as an extensive depth estimation method, showing promising potential for application on tiny robots. Even with bigger sensors (Arducam STARVIS 2 or Nikon D850), the state-of-the-art depth models such as DepthPro \cite{depthPro}, and MiDas \cite{midas} do not generalize to night scenes due to a lack of active illumination. With flashlights (similar to headlights of a vehicle ($\sim 60-80 W$)), these depth models are more accurate but consume a lot more power and can still fail when looking at spurious light sources like street lights. Our approach on the other hand, learns the ``concept'' of structured light with optical defocus and generalizes through these adversaries. 

Our \textit{AsterNet} model was trained \textit{only} with synthesized data and yet is still agonistic to relative placement of the structured light projector and the camera (Fig. \ref{fig:realsense}), because \textit{AsterNet} learned the inherent physics and concept of structured light rather than overfitting to a single extrinsic pair. This leads to repairability as extrinsic alignment is not critical, bringing overall system and repair costs down. \textcolor{black}{We remark that using real data for through online training would improve the depth accuracy further and we leave this for future work.}
\textcolor{black}{Furthermore, on-the-fly adaptation to changes in focal length of the drone camera lens with minimal data would be an important next step for building a unified parsimonious model for passive computation-based depth estimation.
A simple solution would to be to input the focus distance into the network and train with varied focal distances. However, the algorithm for changing focus for navigation is a research topic on its own which we hope to explore in future work. Such an approach would improve the depth prediction accuracy and navigational success rate.
}

Interestingly, all our failure cases (crash into an obstacle/floor or manual override to prevent damage) in flight experiments were attributed to failures of the lower-level Arducopter loiter mode due to a lack of robust position estimates or extreme winds, which upset the PID controllers. We remark that performing experiments in the dark is extremely challenging and potentially unsafe for the robot and people around, since any instability can lead to catastrophic acceleration and crashes. Although a downfacing light enabled us to perform experiments in the dark for optical flow loiter, they were far from reliable; most of the time, the robot would not arm or drift to one side aggressively in the wind due to bad optical flow readings. We believe an active lighting setup for pose estimation would be critical for large-scale and exhaustive tests in multiple environments, and we believe this will be a valuable future contribution to the field. Since obtaining larger sensors with high low-light sensitivity is expensive and difficult, we believe leveraging event cameras for the navigational task would be a next step to help advance navigation in the dark.




\section{Conclusion}
\label{sec:conclusion}
In this work, we presented \textit{AsterNav}, the first navigation system with a monocular camera to enable autonomous quadrotor flight in absolute darkness. By combining an IR camera with structured lighting and a large-aperture coded lens, we showed that depth-dependent blur patterns can be utilized for efficient onboard metric depth estimation on a resource-constrained NVIDIA Jetson Orin$^\text{TM}$ Nano. Our \textit{AsterNet} model, trained entirely on synthetic data generated from PSFs at multiple depths, transferred zero-shot to the real world and enabled autonomous traversals in complex indoor and outdoor environments. Integrated with a simple potential field planner, our system achieved an overall $95.5\%$ navigation success rate, demonstrating both efficiency and robustness. These results highlight the promise of co-designing passive optical elements with perception algorithms for resource-limited aerial robots. This leads to a strong potential for real-world applications, such as search-and-rescue and disaster response, in light-deprived environments. A natural extension of this work is to leverage event cameras, which, owing to their higher dynamic range and temporal resolution, are better suited for high-speed flight and operation in dark or low-light scenes.



\bibliographystyle{IEEEtran}

\bibliography{ref}

\end{document}